\documentclass[sigconf]{acmart}

\AtBeginDocument{%
  }

\usepackage{multirow}
\usepackage[normalem]{ulem}
\useunder{\uline}{\ul}{}
\usepackage{xcolor}
\newcommand\wu[1]{\textcolor{black}{#1}}

\renewcommand\footnotetextcopyrightpermission[1]{} 
\setcopyright{none}
\renewcommand{\shortauthors}{} 

\usepackage{multirow}
\usepackage{multicol}
\usepackage{bm}
\usepackage{amsfonts}

\begin{document}

\title[]{ST-Hyper: Learning High-Order Dependencies Across Multiple Spatial-Temporal Scales for Multivariate Time Series Forecasting}

\author{Binqing Wu}
\authornote{Equal contribution.}
\affiliation{%
  \institution{College of Computer Science and Technology, \\ Zhejiang University}
  \city{Hangzhou}
  \state{Zhejiang}
  \country{China}
}
\email{binqingwu@cs.zju.edu.cn}

\author{Jianlong Huang}
\authornotemark[1]
\affiliation{%
  \institution{College of Computer Science and Technology, \\ Zhejiang University}
  \city{Hangzhou}
  \state{Zhejiang}
  \country{China}
}
\email{22251226@cs.zju.edu.cn}

\author{Zongjiang Shang}
\affiliation{%
  \institution{College of Computer Science and Technology, \\ Zhejiang University}
  \city{Hangzhou}
  \state{Zhejiang}
  \country{China}
}
\email{zongjiangshang@cs.zju.edu.cn}

\author{Ling Chen}
\authornote{Corresponding author.}
\affiliation{%
  \institution{College of Computer Science and Technology, \\ Zhejiang University}
  \city{Hangzhou}
  \state{Zhejiang}
  \country{China}
}
\email{lingchen@cs.zju.edu.cn}

\renewcommand{\shortauthors}{}
\settopmatter{printacmref=false}
\renewcommand\footnotetextcopyrightpermission[1]{}
\setcopyright{none}

\makeatletter
\let\ps@firstpagestyle\ps@empty
\let\ps@standardpagestyle\ps@empty
\makeatother
\pagestyle{empty}

\begin{abstract}
In multivariate time series (MTS) forecasting, many deep learning based methods have been proposed for modeling dependencies at multiple spatial (inter-variate) or temporal (intra-variate) scales. However, existing methods may fail to model dependencies across multiple spatial-temporal scales (ST-scales, i.e.,  scales that jointly consider spatial and temporal scopes). In this work, we propose ST-Hyper to model the high-order dependencies across multiple ST-scales through adaptive hypergraph modeling. Specifically, we introduce a Spatial-Temporal Pyramid Modeling (STPM) module to extract features at multiple ST-scales. Furthermore, we introduce an Adaptive Hypergraph Modeling (AHM) module that learns a sparse hypergraph to capture robust high-order dependencies among features. In addition, we interact with these features through tri-phase hypergraph propagation, which can comprehensively capture multi-scale spatial-temporal dynamics.
Experimental results on six real-world MTS datasets demonstrate that ST-Hyper achieves the state-of-the-art performance, outperforming the best baselines with an average MAE reduction of 3.8\% and 6.8\% for long-term and short-term forecasting, respectively.
\end{abstract}


\maketitle

\section{Introduction}

Multivariate time series (MTS) forecasting plays a critical role in various real-world applications, e.g., transportation management \cite{MTGNN}, environment monitoring\cite{airformer}, and weather prediction\cite{corrformer}. \wu{The task is inherently challenging, as the complex spatial (inter-variable) and temporal (intra-variable) dependencies in MTS should be effectively modeled.}

\wu{Recent deep learning methods have demonstrated strong potential in modeling spatial and temporal dependencies. 
To model spatial dependencies, some works have leveraged graph neural networks (GNNs) \cite{wu2023dstcgcn,yi2024fouriergnn}, hypergraph neural networks (HGNNs) \cite{DHSL,dyhsl}, and Transformer-based architectures \cite{crossformer,itransformer}.
To model temporal dependencies, some works have employed a variety of architectures, including multi-layer perceptrons (MLPs) \cite{Dlinear,wang2025filterts}, recurrent neural networks (RNNs) \cite{lstnet,huang2024itrendrnn}, temporal convolutional networks (TCNs) \cite{tcn,2022timesnet}, and self-attention mechanisms \cite{zhou2022fedformer,piao2024fredformer}.
Despite these advances, these methods only model dependencies at a single spatial or temporal scale, which restricts their capacity to capture the multi-scale dynamics that are often present in real-world time series data.}

\wu{In reality, dependencies at multiple spatial or temporal scales are ubiquitous. For example, weather data often exhibits dependencies at different spatial scales (e.g., city-scale and country-scale) and dependencies at different temporal scales (e.g., day-scale and season-scale).
To capture these dependencies, many methods have been proposed. 
Some methods focus on modeling dependencies at multiple spatial scales \cite{hgcn,GAGNN,MC-STGCN,wu2024weathergnn}. They cluster variables into groups of varying sizes and model variable-level and group-level spatial patterns.  
Some methods focus on modeling dependencies at different temporal scales \cite{2023timemixer, lincyclenet, AdaMSHyper}. They segment time series with different temporal durations or resolutions and model short-term and long-term temporal patterns.
More recently, several methods have attempted to model dependencies across multiple spatial scales and multiple temporal scales independently, either in parallel or sequentially \cite{airformer, MAGNN, cai2024msgnet}. However, these works treat multiple spatial and temporal scales in isolation, potentially overlooking dependencies that arise across \textbf{multiple spatial-temporal scales (ST-scales)}—scales that jointly consider spatial and temporal scopes (e.g., city-day, city-season, and country-season scales in weather data).}

\wu{ST-scales are essential, as some important patterns emerge only at specific ST-scales. For example, the sea-land breeze is a daily phenomenon that occurs in coastal cities, corresponding to a small-spatial-short-temporal scale.
In addition, the Meiyu season, characterized by persistent rainfall, is a seasonal phenomenon observed in cities within the Yangtze River basin, corresponding to a small-spatial-long-temporal scale.
Moreover, climate change (e.g., multi-decade trends across countries) becomes apparent only when analyzed on a large-spatial-long-temporal scale. These examples highlight the necessity of explicitly modeling dependencies across multiple ST-scales to capture the full range of spatial-temporal dynamics of MTS.}

\wu{Despite their importance, modeling such dependencies remains non-trivial. These dependencies are often highly heterogeneous and context-dependent, lacking consistent or pre-defined structural patterns. This variability severely limits the ability to impose unified inductive biases or manually design generalizable structures. 
In addition, dependencies across multiple spatial-temporal scales give rise to intricate cross-scale interactions. Such interactions are not limited to simple pairwise patterns; instead, they often involve high-order patterns that conventional modeling frameworks are fundamentally ill-equipped to capture.}

To this end, we propose ST-Hyper, \wu{which is the first work to incorporate adaptive hypergraph modeling to learn the high-order dependencies across multiple ST-scales for MTS forecasting}. The main contributions of this work are as follows:

\begin{itemize}
\item We introduce a Spatial-Temporal Pyramid Modeling (STPM) module that first learns a spatial pyramidal graph to obtain series at multiple spatial scales, and then employs temporal multi-scale networks and ST-Encoders to extract features across multiple ST-scales. \wu{In particular, we introduce a graph pooling loss for spatial pyramidal graphs based on Laplacian preservation and entropy regularization to encourage the grouping of correlated variables and non-overlapping variable assignments, which can reduce information redundancy.}

\item We introduce an Adaptive Hypergraph Modeling (AHM) module that \wu{learns a \wu{sparse} hypergraph structure to focus on the most correlated features, which can capture the robust high-order dependencies across multiple ST-scales. In addition, we perform tri-phase hypergraph propagation to interact features across multiple ST-scales, which can allow them to reinforce each other and mitigate the impact of anomalies or perturbations. Thereby, AHM can comprehensively and reliably capture multi-scale spatial-temporal dynamics.}

\item We evaluate ST-Hyper on six real-world MTS datasets. The experimental results demonstrate that ST-Hyper achieves state-of-the-art (SOTA) performance, outperforming the best baseline with an average MAE reduction of 3.8\% and 6.8\% for long-term and short-term forecasting, respectively.
\end{itemize}

\section{Related Works}

Due to the superiority in modeling spatial and temporal dependencies, deep-learning-based methods have become the mainstream choices for MTS forecasting. For modeling temporal dependencies, existing methods use MLPs, CNNs, RNNs, and self-attention networks. For example, DLinear \cite{Dlinear} employs two MLPs to model seasonal and trend-cycle components, thereby enabling MLPs to capture trends in time series. LSTNet \cite{lstnet} employs the long short-term memory (LSTM) to model temporal dependencies, but it suffers the problem of gradient vanishing/exploding in RNNs. PatchTST \cite{2022PatchTST} segments time series into multiple subseries, thereby reducing the computational complexity by subseries-level self-attention modeling. For modeling spatial dependencies, existing methods use GNNs, HGNNs, and Transformers. For example, MTGNN \cite{MTGNN} employs a graph structure learning method based on learnable node embeddings to capture dependencies among variables. DyHSL \cite{dyhsl} employs a hypergraph structure learning module to model the high-order dependencies among variables for traffic forecasting.
\wu{MixRNN+ \cite{9956738} captures pairwise and high-order spatial dependencies via hypergraph structure learning for spatial-correlated time series forecasting.}
Crossformer \cite{crossformer} integrates a multi-head self-attention network with learnable vectors as routers to distribute messages across all variables. \wu{Nethertheless, these methods only model dependencies at a single spatial or temporal scale, which restricts their capacity to capture the multi-scale dynamics.}


\wu{To solve this problem, many methods have been proposed to model dependencies at multiple spatial scales or multiple temporal scales.} Some methods aim to model dependencies at multiple spatial scales. For example, HGCN \cite{hgcn} generates hierarchical spatial graphs via spectral clustering on the distance adjacency matrix and uses spatial-temporal graph neural networks (STGNNs) to model spatial and temporal dependencies at each spatial scale. GAGNN \cite{GAGNN} learns a probabilistic assignment matrix between cities and city groups to model spatial dependencies at each spatial scale. 
\wu{HSTGL \cite{lin2024hierarchical} learns micro and macro graphs that are constructed based on transportation connectivity and the geographical correlation, respectively, to model dependencies at two spatial scales.}
Some methods aim to model dependencies at multiple temporal scales. For example, \wu{MSHyper \cite{shang2024mshyper} extracts features at multiple temporal scales and pre-defines high-order dependencies between these features. TimeMixer \cite{2023timemixer} decomposes time series at each temporal scale into seasonal and trend components, then separately mixes multi-scale seasonal and trend representations.} MSGNet \cite{cai2024msgnet} extracts series at different temporal scales based on periods of time series, and employs GNNs and self-attention networks to model spatial and temporal dependencies at each temporal scale, respectively. In addition, some methods model dependencies across multiple spatial scales and dependencies across multiple temporal scales. For example, AirFormer \cite{airformer} leverages two types of self-attention networks to model dependencies across multiple spatial scales and dependencies at each temporal scale. Corrformer \cite{corrformer} employs a cross-correlation mechanism to model dependencies across multiple spatial scales and an auto-correlation mechanism to model dependencies across multiple temporal scales. \wu{However, these methods treat spatial and temporal scales in isolation, potentially overlooking dependencies that arise from their joint interaction.}

\wu{To address this issue, ST-Hyper first extracts features at multiple ST-scales and learn a sparse hypergraph between them adaptively, which can capture robust high-order dependencies across multiple ST-scales. 
Moreover, we conduct tri-phase hypergraph propagation to interact with these features, which can capture multi-scale spatial-temporal dynamics comprehensively and reliably.}

\section{Preliminaries}

\begin{figure*}[ht]
\centering
\includegraphics[width=0.9\textwidth]{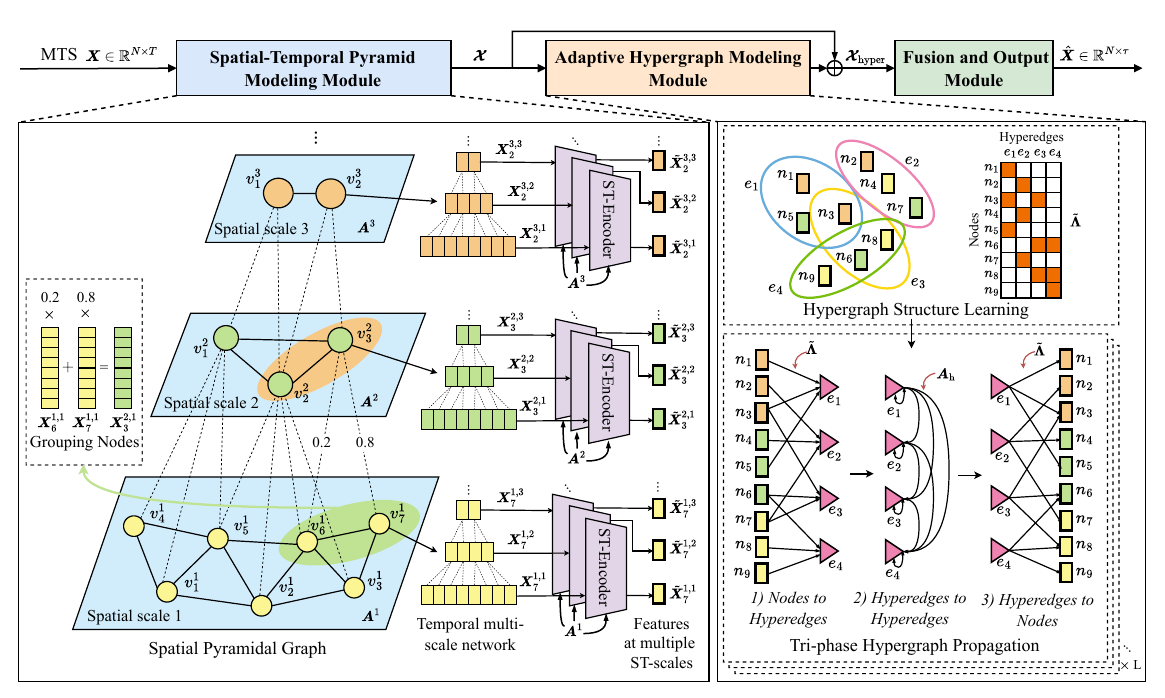}
\caption{The Framework of ST-Hyper.}
\label{figure_framework}
\end{figure*}

\noindent \textbf{MTS Forecasting Formulation.}
MTS forecasting aims to predict future values based on historical values. Formally, we represent historical values as ${\bm X}=\left\{{\bm X}^{1},{\bm X}^{2}, \ldots, {\bm X}^{T}\right\} \in \mathbb{R}^{N \times T}$, where ${N}$ is the number of variables and ${T}$ is the input length of MTS. MTS forecasting is formulated as learning a function $\mathcal{F}$ that maps historical ${T}$ steps of values to future ${\tau}$ steps of values:
\begin{equation}
    \hat{\bm {X}}^{T+1: T+\tau}=\mathcal{F}\left({\bm X} ; \bm{\Theta}\right) \in \mathbb{R}^{N \times \tau},
\end{equation}
where ${\bm \Theta}$ denotes all learnable parameters of $\mathcal{F}$. 

\noindent \textbf{Definition of a Graph for MTS.}
A graph for MTS is defined as $G=(V, E, \bm A)$, where ${V}$ denotes a set of ${N}$ nodes. Each node represents a variable. $v_{i}$ denotes node $i$ in $V$. $E$ denotes a set of edges. Each edge represents a correlation between variables. ${\bm A} \in \mathbb{R}^{N \times N}$ is a weighted adjacency matrix to represent the structure of graph $G$. 

\noindent \textbf{Definition of a Hypergraph for MTS.}
A hypergraph for MTS is defined as $\mathcal{G}=(\mathcal{V}, \mathcal{E})$, where $\mathcal{V}$ denotes a set of nodes, and $\mathcal{E}$ denotes a set of hyperedges. Unlike an edge that only connects two nodes in a graph, a hyperedge in a hypergraph can connect one or more nodes. The hypergraph structure can be represented by an incidence matrix ${\bm {\Lambda} \in \mathbb{R}^{|\mathcal{V}| \times|\mathcal{E}|}}$, where each entry ${\bm {\Lambda}(v,e)}$ indicates whether a node is in a hyperedge or not. A weighted incidence matrix can be formulated as:
\begin{equation}
\bm {\Lambda}(v, e)= \begin{cases} w(v,e), & \text {if} \quad v \in e, \\ 0, & \text {otherwise,}\end{cases}
\end{equation}
where $w(v,e)$ ranges from 0 to 1, representing the assignment possibility from node ${v}$ to hyperedge ${e}$.

\noindent \textbf{Memory-Network-Based Graph Structure Learning.}
A memory network contains an array of memory items (i.e., learnable vectors) to store feature information for pattern matching \cite{weston2014memory}. The memory-network-based graph structure learning method \cite{MegaCRN} augments the generation of node embeddings representing node information in a graph, which can be formulated as:
\begin{equation} \label{graphlearning}
\begin{array}{c}
\boldsymbol{A}=\operatorname{SoftMax}\left(\operatorname{ReLU}\left(\boldsymbol{E}_{1}\left(\boldsymbol{E}_{2}\right)^{\text {T}}\right)\right),\\
\boldsymbol{E}_{1}=\boldsymbol{W}_{\mathrm{E_1}} \boldsymbol{M}, \hspace{1em}
\boldsymbol{E}_{2}=\boldsymbol{W}_{\mathrm{E_2}} \boldsymbol{M},
\end{array}
\end{equation}
where $\bm{A} \in \mathbb{R}^{N \times N}$ denotes the learned adjacency matrix, ${\bm {E}_1}, {\bm E_2} \in \mathbb{R}^{N \times d} $ denote parameterized node embeddings with $ d $ dimensions. $\bm M\in \mathbb{R}^{m \times d}$ denotes $m$ memory items with $d$ dimensions. $\bm W_{\text E_1}, \bm W_{\text E_2} \in \mathbb{R}^{N \times m}$ denote projection matrices that map memory items to node embeddings. 

\section{Methodology}
\subsection{Model Overview}
Empowered by adaptive hypergraph modeling, ST-Hyper aims to model the high-order dependencies across multiple ST-scales. As shown in Figure \ref{figure_framework}, Spatial-Temporal Pyramid Modeling (STPM) module is introduced to first transform input MTS into series at multiple spatial scales and then extract features at each ST-scale. Next, Adaptive Hypergraph Modeling (AHM) module is introduced to model the high-order dependencies across multiple ST-scales. Finally, the features from STPM and AHM modules are aggregated and fed into Fusion and Output module for final predictions.

\subsection{Spatial-Temporal Pyramid Modeling Module} \label{sec:STPM}
STPM module is designed to extract features at multiple ST-scales from fine to coarse. Specifically, STPM incorporates \textit{Spatial Pyramidal Graph (SPG) Learning} to construct graphs at multiple spatial scales and learn mappings between spatial scales, as well as \textit{Multi-Scale Feature Extraction} to transform input MTS into features at multiple ST-scales.

\noindent \textbf{Spatial Pyramidal Graph Learning.}
SPG learning is introduced to first construct graphs at multiple spatial scales. We utilize a memory-network-based graph structure learning method (referred to Equation \ref{graphlearning}) to learn a graph structure for each spatial scale. For the spatial scale $j$, the learned spatial graph can be denoted as $\bm A^{j} \in \mathbb{R}^{N_j \times N_j}$, where $N_{j} = \left\lfloor{{N_{j-1}}/{q}}\right\rfloor$ represents the number of nodes at spatial scale $j$. $q$ is the graph pooling ratio used to coarsen spatial nodes. As shown in Figure \ref{figure_framework}, $v_{i}^{j}$ denotes node $i$ in $\bm A^{j}$. To discover the mapping between two adjacent spatial scales, we learn a probabilistic matrix $\bm{S}^{j} \in \mathbb{R}^ {N_j \times N_{j+1}}$ to assign the nodes at spatial scale $j$ to group nodes at spatial scale $j+1$. The entries of $\bm{S}^{j}$ range from 0 to 1. 

To learn group-wise dependencies between nodes, ST-Hyper integrates a \textbf{graph pooling loss} $\mathcal{L}_{\text{GP}}$ as a regularization term:
\begin{equation} \label{eq_GP}
    \mathcal{L}_{\text{GP}} = \mathcal{L}_{\text{LP}} + \mathcal{L}_{\text E} = || \bm A^{j}_{\text {dtw}},{\bm{S}^{j}} {(\bm{S}^{j})^{\text T}}||_F + \frac1n\sum_{i=1}^n \mathcal{P}(\bm{S}^{j}_{i}),
\end{equation}
where $\mathcal{L}_{\text{LP}}$ represents the Frobenius norm, which intuitively means that nodes with stronger correlations (i.e., larger weights in $\bm A^{j}_{\text {dtw}}$) should be grouped together. $\mathcal{L}_{\text E}$ represents the average entropy loss of all $n$ rows of ${\bm{S}^{j}}$, which makes each row of ${\bm{S}^{j}}$ close to a one-hot vector, so that the membership of each group node can be clearly defined. $\mathcal{P}$ denotes the entropy function. Dynamic Time Warping (DTW) \cite{dtw} can measure the similarity between time series. In equation \ref{eq_GP}, $\bm A^{1}_{\text{dtw}}  \in \mathbb{R}^{N \times N}$ is obtained by computing pairwise DTW distances of $N$ time series over the entire training time range. Then, we can get DTW distance-based adjacency matrix at larger spatial scales based on assignment matrices:
\begin{equation}\label{A_dtw}
    \bm A^{j}_{\text{dtw}} = {(\bm{S}^{j-1})^{\text T}}  \bm A^{j-1}_{\text{dtw}} {\bm{S}^{j-1 }},
\end{equation}
where $\bm A^{j}_{\text{dtw}}  \in \mathbb{R}^{N_j \times N_j}$ represents the DTW distance-based adjacency matrix at spatial scale $j$.

\noindent \textbf{Multi-Scale Feature Extraction.}
This sub-module aims to extract features at multiple ST-scales based on the input MTS $\bm{X} \in \mathbb{R}^{N \times T}$ and the constructed SPG. We define $\bm{X}^{j,k}_{i}$ as features of node $i$ at spatial scale $j$ and temporal scale $k$. The input MTS $\bm{X}$ is represented as $\bm{X}^{1,1} \in \mathbb{R}^{N_{1} \times T_{1}}$, where $N_{1}=N$ and $T_{1}=T$, corresponding to features at the smallest ST-scale. Given $\bm{X}^{1,1}$, we first utilize multiple assignment matrices $\{\boldsymbol{S}^{j}\}_{j=1}^{J-1}$ to obtain series at multiple spatial scales:
\begin{equation}
\label{s_features}
 \bm{X}^{j+1, 1} = (\boldsymbol{S}^{j})^{\text T} \bm{X}^{j, 1}, 
\end{equation}
where $\bm{X}^{j, 1}\in  \mathbb{R}^{N_{j} \times T_{1}}$ represents the series at spatial scale $j$. Thus, we can obtain $  \{\bm{X}^{j,1}\}_{j=1}^{J} \in \mathbb{R}^{( N_{1} +  \ldots + N_{j} + \ldots + N_{J}) \times T_{1}}$, representing series at total $J$ spatial scales. 

To extract features at multiple temporal scales for each spatial scale, we utilize temporal multi-scale networks incorporating 1D CNNs to coarsen the temporal information. Given series at spatial scale $j$ $\bm{X}^{j, 1}$, we employ 1D convolutional layers with kernel size 1$\times${1} and 1$\times 2$ average pooling layers to obtain series at larger temporal scales: 
\begin{equation}
\label{4}
{\bm X}^{j,k} =\mathrm{Pooling}\left(\mathrm{Conv}\left({\bm X}^{j,k-1}\right)\right),
\end{equation}
where ${\bm X}^{j,k}\in \mathbb{R}^{ N_{j} \times T_k}$ represents the series at spatial scale $j$ and temporal scale $k$, with $T_k$ equal to $\frac{T_{1}}{2^{(k-1)}}$. To aggregate information of adjacent time steps, we patchify ${\bm X}^{j,k}$ by dividing each time series into a set of non-overlapping patches \cite{2022PatchTST}:
\begin{equation}
\label{4}
{\bm X}^{j,k}_{\text{patch}} =\mathrm{Linear}\left(\mathrm{Patchify}\left({\bm X}^{j,k}\right)\right),
\end{equation}
where $\bm{X}^{j,k}_{\text {patch}} \in \mathbb{R}^{N_j \times \lfloor{\frac{T_k}{r}}\rfloor \times D} $ represents features at spatial scale $j$ and temporal scale $k$ after patchifying and a linear projection, $r$ is the patch length, and $D$ is the feature dimension. After this process, we can obtain $\boldsymbol{\mathcal{X}} = \{\bm{X}^{j,k}_{\text{patch}}\}_{j=1, k=1}^{J,K} $, representing features at total $J$ spatial scales and $K$ temporal scales (i.e., $J \times K$ ST-scales).

Then, ST-Encoders based on graph convolutional recurrent units (GCRUs) \cite{agcrn} are introduced to extract spatial-temporal information by coupling GNNs and gated recurrent units (GRUs). For each ST-scale, we employ an ST-Encoder with non-shared learnable parameters to extract scale-specific spatial-temporal information: 
\begin{equation} \label{e8}
{\bm{X}}^{j, k}_{\text {en}}= \mathrm{GCRUs}(  {\bm{X}}^{j, k}_{\text{patch}}, \bm A^{j}),
\end{equation} 
where $ {\bm{X}^{j, k}_{\text{en}}}\in \mathbb{R}^{ N 
 \times D}$ represents the features at spatial scale $j$ and temporal scale $k$ after encoding, which is the final hidden state outputted by GCRUs and $D$ is the feature dimension. Recall that the spatial graph at each spatial scale is learned based on a memory network, which can store typical features specific to a spatial scale. Thus, we augment $ {\bm{X}^{j, k}_{\text{en}}}$ by matching it with the memory network at a specific spatial scale: 
\begin{equation}\label{eq_pattern_matching}
\begin{aligned}
\bm{Q}^{j, k} &={\bm{X}^{j, k}_{\text{en}}}{\bm{W}}_{j}+\bm{b}_{j},
\\
{\bm{X}}^{j, k}_{\text{att}} &=  \mathrm{SoftMax}(\bm{Q}^{j,k}({\bm{M}^{j}})^{\text T}) \bm{M}^{j}, 
\\
\tilde{\bm{X}}^{j, k} &= \mathrm{Concat}({\bm{X}}^{j, k}_{\text{att}}, {\bm{X}^{j, k}_{\text{en}}}),    
\end{aligned}
\end{equation}
where $\bm{Q}^{j, k} \in \mathbb{R}^{N_{j} \times d}$ denotes the query matrix. $\bm {M}^{{j}}\in \mathbb{R}^{ m \times d}$ represents the $m$ memory items with $d$ dimensions stored in the memory network at spatial scale $j$. ${\bm{X}}^{j, k}_{\text{att}}  \in \mathbb{R}^{N_j \times d}$ denotes the features after an attention-based pattern matching operation. After that, we concatenate ${\bm{X}}^{j, k}_{\text{att}}$ and ${\bm{X}^{j, k}_{\text{en}}}$ to obtain the features augmented by a memory network $\tilde{\bm{X}}^{j,k} \in \mathbb{R}^{ N_{j} 
 \times (D+d)}$ . Through STPM module, we can obtain ${\bm{\mathcal{X}}} = \{\tilde{\bm{X}}^{j,k}\}_{j=1, k=1}^{J,K} \in \mathbb{R}^{(N_1+\ldots+N_{J}) \times K \times (D+d)}$, representing features at $J \times K$ ST-scales.

\subsection{Adaptive Hypergraph Modeling Module}
AHM module is introduced to model the high-order dependencies across multiple ST-scales through adaptive hypergraph modeling. Specifically, AHM incorporates \textit{hypergraph structure learning} to discover group-wise dependencies among features from multiple ST-scales and \textit{tri-phase hypergraph propagation} to further capture the high-order dependencies across multiple ST-scales.

\noindent \textbf{Hypergraph Structure Learning.}
We treat each feature of ${\bm{\mathcal{X}}} $ as a node in a hypergraph. Thus, the total number of nodes in a hypergraph $\alpha$ is $(N_{1}+\ldots+N_{J}) \times K$. The information of each node in a hypergraph comes from the corresponding feature of  ${\bm{\mathcal{X}}} $. We introduce a randomly initialized weighted incidence matrix ${\bm {\Lambda}} \in \mathbb{R}^{\alpha \times \beta}$ to reflect the hypergraph structure, where $\beta$ denotes the number of hyperedges. Each entry of ${\bm {\Lambda}}$ (ranging from 0 to 1) represents the probability of assigning a node to a hyperedge. However, learning a dense incidence matrix will result in too many nodes being associated with each hyperedge. To this end, each column of ${\bm {\Lambda}}$ is sparsified by selecting top $K^\prime$ weights, indicating that the most correlated $K^\prime$ nodes are associated with a hyperedge. Thus, we can get the sparsified incidence matrix $\tilde {\bm {\Lambda}} \in \mathbb{R}^{\alpha \times \beta}$, which can discover the group-wise dependencies among nodes.

\noindent \textbf{Tri-phase Hypergraph Propagation.}
Tri-phase hypergraph propagation is introduced to model the interaction between nodes and hyperedges based on the learned hypergraph structure, consisting of nodes to hyperedges, hyperedges to hyperedges, and hyperedges to nodes propagation phases. Compared to the traditional hypergraph convolution \cite{hypergraphbai}, there are two advantages: 1) Hyperedges to hyperedges phase can enhance the dependency modeling among hyperedges by learning a hyperedge graph and employing graph attention networks (GATs) to adaptively interact hyperedges. 2) Hyperedges to nodes phase can update node information by adaptively interacting nodes with their associated hyperedges based on an attention mechanism.

\textit{1) Nodes to Hyperedges.} This phase aims to make each hyperedge aggregate the information from nodes associated with it, so that each hyperedge can represent a group of features across multiple ST-scales. This process can be formulated as:
\begin{equation}
    {\bm{\mathcal E}_{1}} = \mathrm{\phi}(\bm{U} \tilde{\bm {\Lambda}}^{{\text T}} {\bm{\mathcal{X}}} ) + \tilde{\bm {\Lambda}}^{{\text T}} {\bm{\mathcal{X}}}, 
\end{equation}
where $ {\bm{\mathcal E}_{1}} \in \mathbb{R}^{\beta \times {D}_{\text e}}$ represents the obtained hyperedges and $D_{\text e} = D+d$ is the hyperedge dimension. $\bm{U} \in \mathbb{R}^{\beta \times \beta}$ represents the normalized weights of hyperedges. ${\phi}$ denotes an activation function.

\textit{2) Hyperedges to Hyperedges.} Since each hyperedge is aggregated from a group of nodes, modeling the interaction among hyperedges is significant for capturing group-wise dependencies. To discover the relationships between hyperedges, we utilize the memory-network-based graph learning method (referred to Equation \ref{graphlearning}) to learn a hyperedge graph ${\bm{A}}_{\text h} \in \mathbb{R}^{\beta \times \beta }$. Moreover, we leverage a GAT \cite{GAT2017} to adaptively determine the importance of neighboring hyperedges to each hyperedge:
\begin{gather}
{\bm{\mathcal E}_{2}} = \mathrm{GAT}({\bm{\mathcal E}_{1}}, \bm{A}_{{\text h}}),
\end{gather}
where ${\bm{\mathcal E}_{2}} \in \mathbb{R}^{\beta \times {D}_{\text e}}$ denotes the updated hyperedges. The memory network can store typical features shared on the hyperedge graph. Similar to Equation \ref{eq_pattern_matching}, we augment ${\bm{\mathcal E}_{2}}$ by an attention-based pattern matching operation:
\begin{equation}
\begin{aligned}
\bm{Q_{\text h}} &={\bm{\mathcal E}_{2}}{\bm{W}}_{\text Q }+\bm{b}_{\text Q},
\\
{\bm{\mathcal E}}_{2}^{\prime} &=  \mathrm{SoftMax}(\bm{Q}_{\text h}({\bm{M}_{\text h}})^{\text {T}}) {\bm{M}_{\text h}}, 
\\
\tilde{\bm{\mathcal E}_{2}} &= \mathrm{Concat({\bm{\mathcal E}_{2}}, {\bm{\mathcal E}}_{2}^{\prime})},    
\end{aligned}
\end{equation}
where $\bm{Q}_{\text h} \in \mathbb{R}^{\beta \times d}$ denotes the query matrix. ${\bm{\mathcal E}}_{2}^{\prime} \in \mathbb{R}^{\beta \times d}$ denotes the hyperedges after interacting with the memory items $\bm M_{\text h} \in \mathbb{R}^{m \times d}$. We can get memory-augmented hyperedges $ \tilde{\bm{\mathcal E}_{2}} \in \mathbb{R}^{\beta \times ({D}_{\text e} +d)}$ by concatenating ${\bm{\mathcal E}_{2}}$ and ${\bm{\mathcal E}}_{2}^{\prime}$. Furthermore, we aggregate ${\bm{\mathcal E}_{1}}$ and $ \tilde{\bm{\mathcal E}_{2}}$ by an MLP:
\begin{equation}
       {\hat{\bm{\mathcal E}}} = 
     \mathrm{MLP}({\bm{\mathcal E}_{1}} + (\bm{W} \tilde{\bm{\mathcal E}_{2}} + \bm{b})),
\end{equation}
where ${\hat{\bm{\mathcal E}}} \in \mathbb{R}^{\beta \times {D}_{\text e}}$ represents the updated hyperedges.

\textit{3) Hyperedges to Nodes.} After the propagation between hyperedges, we update node information by interacting nodes with their associated hyperedges based on an attention mechanism. According to the incidence matrix $ \tilde{\bm {\Lambda}}$, we can get an attention mask to identify which hyperedges are associated with each node:
\begin{equation}
    {\bm {\Gamma}}_{i,j} = \begin{cases}
        0, &\tilde{\bm {\Lambda}}_{i,j} \neq
 0,\\ 
        -\infty, &\tilde{\bm {\Lambda}}_{i,j} = 0.
    \end{cases}
\end{equation}
Based on the attention mask $\bm{\Gamma}$, we employ an attention mechanism to interact nodes with their associated hyperedges: 
\begin{equation}
\renewcommand{\arraystretch}{1.3} 
\begin{array}{c}
\bm{Q }={\bm{\mathcal{X}}}{\bm{W}}_{\text q}+\bm{b}_{\text q},\hspace{1em} \bm{K } = {\hat{\bm{\mathcal E}}}{\bm{W}}_{\text k}+\bm{b}_{\text k}, \hspace{1em}\bm{V }= {\hat{\bm{\mathcal E}}}{\bm{W}}_{\text v}+\bm{b}_{\text v},
\\
{\bm{\mathcal{X}}}_{\text{att}} =  \mathrm{SoftMax}(\bm{Q}{\bm{K}^{\text {T}} +\bm{\Gamma}) {\bm{V}}}, 
\\
{\bm{\mathcal{X}}}_{\text{hyper}} = \mathrm{LayerNorm}(\mathrm{MLP}({\bm{\mathcal{X}}}_{\text{att}} ) + {\bm{\mathcal{X}}}),
\end{array}
\end{equation}
where $\bm{Q}\in \mathbb{R}^{\alpha\times (D + d)}$ represents the query matrix projected from nodes, $\bm{K}, \bm{V}\in \mathbb{R}^{\beta \times D_{\text e}}$ represent the key matrix and the value matrix, respectively, projected from hyperedges, and $D_e = D + d$. ${\bm{\mathcal{X}}}_{\text{att}} \in \mathbb{R}^{\alpha \times D_{\text {e}}}$ represents the attention output after the interaction between nodes and hyperedges. ${\bm{\mathcal{X}}}_{\text{hyper}}  = \{{\bm{X}}^{j,k}_{\text{hyper}}\}_{j=1, k=1}^{J,K} \in \mathbb{R}^{(N_1+\ldots+N_{J}) \times K \times D} $ represents the updated nodes (i.e., the output of AHM module).

\subsection{Fusion and Output Module}
\noindent \textbf{Fusion.}
We fuse features ${\bm{\mathcal{X}}}_{\text{hyper}} $ to get the fused representations for $N$ variables. We first fuse features at all temporal scales for each spatial scale. We use normalized weights $\{\omega^{j, k}\}_{k=1}^{K}$ to decide the contribution of features at each temporal scale and obtain the fused features at spatial scale $j$ ${\bm{\mathcal{O}}}^{j} \in \mathbb{R}^{N_{j} \times D}$. After that, we fuse features at all spatial scales by passing features at larger spatial scales to features at spatial scale 1. These two processes can be formulated as:
\begin{equation}\label{fusion} 
     {\bm{\mathcal{O}}}^{j} = \sum_{k=1}^{K} (\omega^{j,k} {\bm{{X}}}^{j,k}_{\text{hyper}}), \hspace{1em} {\bm{X}}_{\text {fused}} = {\bm{\mathcal{O}}}^{1} + \sum_{j=2}^{J}\prod_{l=0}^{j-1}\bm{S}^{l}{\bm{\mathcal{O}}}^{j},
\end{equation}
where $\prod_{l=0}^{j-1}\bm{S}^{l} \in \mathbb{R}^{N_{1} \times N_{j} }$ represents probabilities to pass features at spatial scale $j$ to features at spatial scale 1. $\bm{X}_{\text {fused}} \in \mathbb{R}^{N_{1} \times D_{\text e}}$ represents the fused representations for $N$ variables, with $N_{1} = N$.

\noindent \textbf{Output.}
MTS forecasting tasks can be divided into short-term and long-term forecasting, which depends on whether output length is longer than input length \cite{2022timesnet}. For short-term forecasting, to enhance the dependency modeling of adjacent time steps, we use GCRUs to decode $\bm{X}_{\text {fused}} \in \mathbb{R}^{N \times D_{\text e}}$ and output the next $\tau$ hidden states $\hat{\bm{H}} \in \mathbb{R}^{N \times \tau \times D_{\text {e}}}$ recurrently. $\hat{\bm{H}}$ is then projected to get $\tau$ predictions $\hat{{\bm{X}}} \in \mathbb{R}^ {N \times \tau}$.  For long-term forecasting, to avoid the cumulative error caused by sequential modeling, we use MLPs to transform $\bm{X}_{\text {fused}} \in \mathbb{R}^{N \times D_{\text e}}$ to $\hat{{\bm{X}}} \in \mathbb{R}^ {N \times \tau}$. We integrate $\mathcal{L}_{1}$ loss and graph pooling loss $\mathcal{L}_{\text{GP}}$ (Eq.4) as the training loss: 
\begin{equation}
    \mathcal{L}_{1} =  \sum_{i=1}^{N}|\hat{{\bm X}}^{T+1:T+\tau}-\bm X^{T+1:T+\tau}|,
    \hspace{1em}
    \mathcal L_{\text {train}} = \mathcal L_{1} + \lambda \mathcal L_{\text {GP}},
\end{equation}
where $\hat{{\bm X}}^{T+1:T+\tau}$ is predicted values and $\bm X^{T+1:T+\tau}$ is ground truth. $\lambda$ is a balancing factor that controls the importance of $\mathcal L_{\text {GP}}$ to the training loss $\mathcal L_{\text {train}}$, which ranges from 0 to 1. 

\section{Experiments} 

\subsection{Experimental Settings} 
\noindent \textbf{Datasets.}
\wu{In light of the number of variables and their inter-variate correlations, we select six public datasets comprising over 100 variables to ensure sufficient complexity and analytical depth.} The statistics of these datasets are summarized in Table \ref{t_dataset}.

METR-LA and PEMS-BAY datasets are provided by MTGNN \cite{MTGNN}. China-AQI dataset is provided by GAGNN \cite{GAGNN}. Electricity and Solar-Energy datasets are provided by Autoformer \cite{autoformer}. Temperature dataset is provided by Corrformer \cite{corrformer}.
\wu{We conduct the pre-processing strategies on datasets according to the original established protocols in the corresponding papers. We split datasets chronologically for training, validation, and testing with the ratio 7:1:2 for all datasets.}

\begin{table}[]
\centering
\caption{Dataset statistics.\label{t_dataset}}

\setlength{\tabcolsep}{2pt} 
\resizebox{\linewidth}{!}{
\begin{tabular}{@{}lcccccc@{}}
\toprule
Datasets     & \# Samples & \# Variables & Sample rate & Input length & Output length     & Feature                \\ \midrule
METR-LA      & 34,272     & 207          & 5 minutes   & 12           & 12                & Traffic speed          \\
PEMS-BAY     & 52,116     & 325          & 5 minutes   & 12           & 12                & Traffic speed          \\
China-AQI    & 20,400     & 209          & 1 hour      & 96           & 24                & Air Quality Index      \\ \midrule
Electricity  & 26,304     & 321          & 1 hour      & 96           & 96, 192, 336, 720 & Electricity consumption \\
Solar-Energy & 52,560     & 137          & 10 minutes  & 96           & 96, 192, 336, 720 & Solar power generation \\ 
 Temperature& 17,544& 3,850& 1 hour& 96& 192&Temperature\\ \bottomrule
\end{tabular}
}

\end{table}

\begin{table}[]
\centering
\caption{Hyperparameter search space and choices on different datasets. }
\label{hyperparameters}

\resizebox{\linewidth}{!}{
\begin{tabular}{l|c|c|c|c|c|c|c}
\toprule
Hyperparameters                                         & \multicolumn{1}{c|}{Search Space} & \multicolumn{1}{l|}{METR-LA} & \multicolumn{1}{l|}{PEMS-BAY} & \multicolumn{1}{l|}{China-AQI} & \multicolumn{1}{l|}{Electricity} & \multicolumn{1}{l|}{Solar-Energy} & \multicolumn{1}{l}{Temperature} \\ \midrule
Spatial graph pooling ratio $q$                         & \{10, 20, 30, 40, 50\}                & 10& 20                            & 20                             & 20                               & 20& 30                              \\ \midrule
Number of spatial scales $J$                      & \{1, 2, 3, 4\}                       & 2                            & 2                             & 2                              & 2                                & 2                                 & 3                               \\ \midrule
Patch length $r$                                        & \{2, 4, 8, 16, 24\}                   & 2                            & 2                             & 16                             & 8                                & 16& 16                              \\ \midrule
Number of temporal scales $K$                     & \{1, 2, 3, 4\}                       & 3                            & 3                             & 3                              & 3                                & 3                                 & 3                               \\ \midrule
Number of $ K'$                                            & \{10, 20, 30, 40, 50\}                & 20                           & 20                            & 20                             & 10                               & 10                                & 20                              \\ \midrule
Balancing factor of $\mathcal{L}_{\text{GP}}$ $\lambda$ & \{1e-3, 5e-2, 1e-2, 5e-1, 1e-1\}      & 1e-1& 1e-1& 1e-2                           & 1e-2                             & 1e-2                              & 1e-2\\ \bottomrule
\end{tabular}
}

\end{table}

\begin{table}[]
\centering
\caption{Results for long-term forecasting.}
\label{tab:long-term}

\resizebox{\linewidth}{!}{
\begin{tabular}{c|c|cccc|cccc|c}
\toprule
\multirow{2}{*}{Method} & Dataset & \multicolumn{4}{c|}{Electricity} & \multicolumn{4}{c|}{Solar-Energy} & Temperature \\
\cmidrule(l){2-11} 
 & Horizon & 96 & 192 & 336 & 720 & 96 & 192 & 336 & 720 & 192 \\
\midrule
ST-Hyper & MSE & \textbf{0.143} & {\ul 0.166} & {\ul 0.181} & \textbf{0.225} & \textbf{0.185} & \textbf{0.212} & \textbf{0.229} & \textbf{0.236} & \textbf{0.267} \\
(Ours) & MAE & \textbf{0.237} & 0.267 & {\ul 0.270} & {\ul 0.316} & \textbf{0.210} & \textbf{0.237} & \textbf{0.249} & \textbf{0.261} & \textbf{0.380} \\ \midrule
MSHyer & MSE & 0.152$^*$ & 0.171$^*$ & 0.187$^*$ & {\ul 0.224}$^*$ & 0.215 & {\ul 0.223} & 0.249 & 0.264 & 0.301 \\
(2024) & MAE & 0.252$^*$ & 0.271$^*$ & 0.284$^*$ & {\ul 0.316}$^*$ & 0.264 & 0.273 & 0.290 & 0.308 & 0.422 \\ \midrule
iTransformer & MSE & 0.148$^*$ & \textbf{0.162}$^*$ & \textbf{0.178}$^*$ & \textbf{0.225}$^*$ & 0.203$^*$ & 0.233$^*$ & 0.248$^*$ & 0.249$^*$ & 0.288 \\
(2024) & MAE & {\ul 0.240}$^*$ & \textbf{0.253}$^*$ & \textbf{0.269}$^*$ & 0.317$^*$ & {\ul 0.237}$^*$ & {\ul 0.261}$^*$ & {\ul 0.273}$^*$ & {\ul 0.275}$^*$ & 0.391 \\ \midrule
TimeMixer & MSE & 0.153$^*$ & {\ul 0.166}$^*$ & 0.185$^*$ & \textbf{0.225}$^*$ & {\ul 0.195} & 0.224 & {\ul 0.237} & {\ul 0.242} & 0.292 \\
(2024) & MAE & 0.247$^*$ & {\ul 0.256}$^*$ & 0.277$^*$ & \textbf{0.310}$^*$ & 0.266 & 0.287 & 0.295 & 0.298 & 0.396 \\ \midrule
MSGNet & MSE & 0.166$^*$ & 0.184$^*$ & 0.195$^*$ & 0.231$^*$ & 0.232 & 0.256 & 0.279 & 0.287 & 0.298 \\
(2024) & MAE & 0.274$^*$ & 0.292$^*$ & 0.302$^*$ & 0.332$^*$ & 0.277 & 0.295 & 0.314 & 0.32 & 0.411 \\ \midrule
Corrformer & MSE & 0.179 & 0.196 & 0.218 & 0.251 & 0.211 & 0.237 & 0.252 & 0.264 & {\ul 0.286} \\
(2023) & MAE & 0.285 & 0.293 & 0.320 & 0.339 & 0.289 & 0.301 & 0.314 & 0.328 & {\ul 0.389} \\ \midrule
AirFormer & MSE & 0.178 & 0.192 & 0.205 & 0.242 & 0.204 & 0.225 & 0.250 & 0.256 & 0.321 \\
(2023) & MAE & 0.281 & 0.290 & 0.309 & 0.334 & 0.276 & 0.293 & 0.303 & 0.306 & 0.434 \\ \midrule
PatchTST & MSE & 0.181$^*$ & 0.188$^*$ & 0.204$^*$ & 0.246$^*$ & 0.234$^*$ & 0.267$^*$ & 0.290$^*$ & 0.289$^*$ & 0.295 \\
(2023) & MAE & 0.270$^*$ & 0.274$^*$ & 0.293$^*$ & 0.324$^*$ & 0.286$^*$ & 0.310$^*$ & 0.315$^*$ & 0.317$^*$ & 0.399 \\ \midrule
Crossformer & MSE & 0.219$^*$ & 0.231$^*$ & 0.246$^*$ & 0.280$^*$ & 0.310$^*$ & 0.734$^*$ & 0.750$^*$ & 0.769$^*$ & 0.335 \\
(2023) & MAE & 0.314$^*$ & 0.322$^*$ & 0.337$^*$ & 0.363$^*$ & 0.331$^*$ & 0.725$^*$ & 0.735$^*$ & 0.765$^*$ & 0.444 \\ \midrule
CrossGNN & MSE & 0.173$^*$ & 0.195$^*$ & 0.206$^*$ & 0.231$^*$ & 0.297 & 0.357 & 0.374 & 0.376 & 0.307 \\
(2023) & MAE & 0.275$^*$ & 0.288$^*$ & 0.300$^*$ & 0.335$^*$ & 0.339 & 0.372 & 0.389 & 0.387 & 0.414 \\ \midrule
GAGNN & MSE & 0.227 & 0.248 & 0.275 & 0.295 & 0.293 & 0.342 & 0.395 & 0.411 & 0.322 \\
(2023) & MAE & 0.329 & 0.369 & 0.383 & 0.374 & 0.326 & 0.354 & 0.402 & 0.429 & 0.423 \\ \midrule
MTGNN & MSE & 0.217$^*$ & 0.238$^*$ & 0.260$^*$ & 0.290$^*$ & 0.337 & 0.360 & 0.377 & 0.399 & 0.325 \\
(2020) & MAE & 0.318$^*$ & 0.352$^*$ & 0.348$^*$ & 0.369$^*$ & 0.353 & 0.380 & 0.398 & 0.422 & 0.431 \\
\bottomrule
\end{tabular}}

\end{table}

\noindent \textbf{Baselines.}
We compare ST-Hyper with SOTA MTS methods for long- and short-term forecasting, including: 
\textbf{MSHyper} \cite{shang2024mshyper} builds multi-scale temporal features via pooling and hypergraphs; 
\textbf{iTransformer} \cite{itransformer} embeds series as variable tokens and models attention among them; 
\textbf{TimeMixer} \cite{2023timemixer} samples series at multiple scales, decomposes them into seasonal and trend parts, and mixes them separately; 
\textbf{MSGNet} \cite{cai2024msgnet} extracts series by periods and models spatial/temporal dependencies with GNNs and self-attention; 
\textbf{Corrformer} \cite{corrformer} uses cross- and auto-correlation for spatial and temporal dependencies; 
\textbf{AirFormer} \cite{airformer} employs dual self-attention for multi-scale spatial and temporal modeling; 
\textbf{PatchTST} \cite{2022PatchTST} segments series into subseries-level patches with channel-independent Transformers; 
\textbf{Crossformer} \cite{crossformer} merges adjacent steps for multi-scale features and uses a learnable router with self-attention; 
\textbf{CrossGNN} \cite{crossGNN} extracts series by periods and models dependencies via cross-scale and cross-variable GNNs; 
\textbf{GAGNN} \cite{GAGNN} constructs geometric graphs and pools them across scales; 
\textbf{MTGNN} \cite{MTGNN} uses 1D CNNs for multi-scale features, stacking GCNs for spatial and TCNs for temporal dependencies.

\wu{Moreover, we compare ST-Hyper with SOTA STGNNs focusing on various graph structure learning (adaptive, hierarchical, high-order), including: 
\textbf{MegaCRN} \cite{MegaCRN} uses learnable embeddings for spatial and temporal heterogeneity; 
\textbf{PDFormer} \cite{pdformer} employs learnable embeddings emphasizing delay effects; 
\textbf{STAFormer} \cite{liu2023spatio} adapts Vanilla Transformer with adaptive embeddings for spatio-temporal dependencies; 
\textbf{HGCN} \cite{hgcn} builds geometric graphs and applies spectral clustering for hierarchical structures; 
\textbf{HiGP} \cite{cini2024graph} models hierarchical dependencies with trainable pooling and differentiable reconciliation; 
\textbf{MixRNN} \cite{liang2022mixed} introduces relation-aware mixers and physics-informed residuals for higher-order spatial dependencies; 
\textbf{ASTHGCN} \cite{zhu2023asthgcn} uses time-varying learnable hypergraphs for dynamic higher-order spatial modeling; 
\textbf{MvHSTM} \cite{shenmvhstm} adopts temporal transformers and dual learnable hypergraphs for local and cross-regional dependencies.}

\noindent \textbf{Training Details.}
We conduct the experiment five times with different random seeds and report the average value for evaluation metrics. Following existing works, we use Mean Absolute Error (MAE), Root Mean Squared Error (RMSE), and Mean Absolute Percentage Error (MAPE) for short-term forecasting \cite{MTGNN}, and use MAE and Mean Squared Error (MSE) for long-term forecasting \cite{itransformer}. Our implementation of ST-Hyper is in Python 3.8, utilizing PyTorch 1.9.0, and trained on NVIDIA GeForce RTX 3090 Ti GPU. 
The code is available at the link \footnote{https://anonymous.4open.science/ST-Hyper-83E7}.
We employ the Adam optimizer with an initial learning rate of 0.001. The model is trained for 150 epochs, and the training is early stopped if the validation loss fails to decrease with 15 epochs. The batch size is set to 8 for Temperature datasets, and 32 for others. The hidden dimension of GCRUs is set to 64. The memory number is set to 20, with dimension 32. The number of layers of AHM module is set to 1. The number of hyperedges is set to 40. The details of hyperparameter tuning for different datasets are given in Table \ref{hyperparameters}. 

\subsection{Main Results}

We evaluate ST-Hyper and baselines on six datasets for long-term and short-term forecasting. \textbf{Bold} and \underline{underlined} indicate the best and second-best performance, respectively. The results are summarized in Table \ref{tab:long-term}, Table \ref{tab:short-term}, and Table \ref{tab:stgnns}. Notably, the results for baselines marked with $^*$ in Table \ref{tab:long-term} are sourced from iTransformer \cite{itransformer} or their original publications. Those with $^*$ in Table \ref{tab:stgnns} are sourced from STAFormer \cite{liu2023spatio} or their original publications.
Other results without marks are obtained by rerunning the official codes using default settings. From these results, we can observe that:

(1) \wu{As shown in Table \ref{tab:long-term}, for long-term forecasting, ST-Hyper outperforms the best baselines with an average MAE reduction of 3.8\%, particularly on Solar-Energy and Temperature datasets with complex spatio-temporal dependencies, highlighting the importance of modeling multi-scale dependencies.}

(2) \wu{As shown in Table \ref{tab:short-term}, for short-term forecasting, ST-Hyper consistently surpasses all baselines, achieving an average MAE reduction of 6.8\%. Its tri-phase hypergraph propagation allows adaptation to local variations while preserving global context, effectively extracting meaningful patterns and improving robustness to noisy spatial-temporal data, i.e.,  traffic speed and air quality.}

\begin{table}[]
\centering
\caption{Results for short-term forecasting.}
\label{tab:short-term}

\resizebox{\linewidth}{!}{
\begin{tabular}{c|c|ccc|ccc|cccc}
\toprule
\multirow{2}{*}{Method} & Dataset & \multicolumn{3}{c|}{METR-LA} & \multicolumn{3}{c|}{PEMS-BAY} & \multicolumn{4}{c}{China-AQI} \\ \cmidrule(l){2-12} 
 & Horizon & 3 & 6 & 12 & 3 & 6 & 12 & 3 & 6 & 12 & 24 \\ \midrule
ST-Hyper & MAE & \textbf{2.49} & \textbf{2.86} & \textbf{3.27} & \textbf{1.28} & \textbf{1.60} & \textbf{1.87} & \textbf{10.88} & \textbf{15.31} & \textbf{19.86} & \textbf{24.24} \\
(Ours) & RMSE & \textbf{4.90} & \textbf{5.95} & \textbf{7.08} & \textbf{2.71} & \textbf{3.68} & \textbf{4.42} & \textbf{17.47} & \textbf{23.07} & \textbf{30.50} & \textbf{38.39} \\
 & MAPE & \textbf{6.34} & \textbf{7.77} & \textbf{9.37} & \textbf{2.68} & \textbf{3.59} & \textbf{4.47} & \textbf{16.92} & \textbf{24.82} & \textbf{33.48} & \textbf{39.8} \\
\midrule
MSHyer & MAE & 3.54 & 4.47 & 6.22 & 1.62 & 2.05 & 2.84 & 12.83 & 18.06 & 23.02 & 26.40 \\
(2024) & RMSE & 6.73 & 8.92 & 10.5 & 3.46 & 4.25 & 6.24 & 22.67 & 30.87 & 36.74 & 43.00 \\
 & MAPE & 8.60 & 11.64 & 12.7 & 3.52 & 4.41 & 6.38 & 21.75 & 29.44 & 36.36 & 44.26 \\
\midrule
iTransformer & MAE & 4.01 & 5.29 & 7.24 & 1.54 & 2.08 & 2.93 & 12.41 & 17.36 & {\ul 22.07} & 26.09 \\
(2024) & RMSE & 9.55 & 12.18 & 15.42 & 3.23 & 4.62 & 6.4 & 21.91 & 29.19 & {\ul 35.83} & 41.19 \\
 & MAPE & 9.00 & 11.71 & 15.96 & 3.19 & 4.48 & 6.63 & 19.22 & 27.21 & 35.48 & 41.5 \\
\midrule
TimeMixer & MAE & 2.88 & 3.22 & 3.87 & 1.61 & 2.14 & 2.97 & 13.96 & 18.37 & 23.84 & 27.57 \\
(2024) & RMSE & 5.88 & 6.27 & 7.91 & 3.33 & 4.74 & 6.60 & 24.44 & 30.37 & 37.19 & 42.25 \\
 & MAPE & {\ul 6.72} & 8.33 & 10.72 & 3.30 & 4.66 & 6.82 & 21.28 & 29.89 & 36.27 & 42.46 \\
\midrule
MSGNet & MAE & 3.72 & 4.35 & 6.2 & 1.52 & 1.86 & 2.42 & 18.45 & 20.77 & 23.58 & 26.24 \\
(2024) & RMSE & 7.10 & 8.71 & 10.91 & 3.40 & 4.27 & 5.51 & 29.46 & 32.23 & 36.02 & 40.84 \\
 & MAPE & 9.24 & 11.33 & 13.48 & 3.19 & 4.26 & 5.37 & 30.09 & 33.15 & 38.04 & 43.93 \\
\midrule
Corrformer & MAE & 3.26 & 3.98 & 4.03 & 1.55 & 1.98 & 2.59 & {\ul 11.74} & {\ul 16.95} & 22.46 & 26.54 \\
(2023) & RMSE & 6.16 & 7.25 & 8.84 & 3.24 & 4.50 & 5.87 & 21.63 & 29.46 & 36.78 & 42.66 \\
 & MAPE & 7.65 & 9.47 & 11.15 & 3.22 & 4.36 & 6.23 & 20.41 & 27.38 & 36.13 & 44.06 \\
\midrule
AirFormer & MAE & 3.6 & 4.24 & 4.93 & 1.46 & 1.93 & 2.53 & 11.87 & 17.1 & 22.12 & 26.18 \\
(2023) & RMSE & 6.84 & 8.38 & 9.05 & 3.19 & 4.43 & 5.81 & {\ul 21.37} & 29.13 & 36.43 & 42.04 \\
 & MAPE & 9.15 & 10.57 & 12.61 & 3.11 & 4.31 & 6.03 & 20.11 & {\ul 26.85} & 35.72 & 43.74 \\
\midrule
PatchTST & MAE & 4.27 & 4.85 & 5.68 & 3.57 & 4.15 & 4.49 & 13.49 & 18.88 & 23.98 & 27.20 \\
(2023) & RMSE & 7.21 & 8.36 & 9.74 & 5.91 & 6.70 & 7.41 & 24.01 & 30.88 & 38.46 & 42.81 \\
 & MAPE & 9.45 & 10.61 & 12.58 & 7.52 & 9.10 & 10.25 & 19.73 & 28.67 & 38.70 & 45.22 \\
\midrule
Crossformer & MAE & 4.65 & 5.35 & 6.49 & 4.76 & 5.47 & 5.96 & 13.08 & 17.82 & 22.85 & {\ul 25.99} \\
(2023) & RMSE & 7.95 & 9.11 & 10.65 & 7.93 & 8.55 & 9.12 & 23.67 & {\ul 28.28} & 36.42 & 40.67 \\
 & MAPE & 9.85 & 11.29 & 13.64 & 11.08 & 13.23 & 14.79 & 19.72 & 28.35 & 39.24 & 45.47 \\
\midrule
CrossGNN & MAE & {\ul 2.68} & {\ul 3.03} & {\ul 3.40} & 1.46 & 1.85 & 2.37 & 12.49 & 17.82 & 23.69 & 29.11 \\
(2023) & RMSE & {\ul 5.18} & {\ul 5.79} & {\ul 7.20} & 3.05 & 4.21 & 5.40 & 22.87 & 30.83 & 38.62 & 46.00 \\
 & MAPE & 6.81 & {\ul 7.98} & 9.91 & 3.03 & 4.10 & 5.67 & 20.91 & 28.94 & 38.84 & 50.45 \\
\midrule
GAGNN & MAE & 4.35 & 4.78 & 5.59 & 1.93 & 2.43 & 2.81 & 12.45 & 17.22 & 22.23 & 26.27 \\
(2023) & RMSE & 7.55 & 8.71 & 10.48 & 3.78 & 4.63 & 5.78 & 22.45 & 30.16 & 37.35 & 43.53 \\
 & MAPE & 9.79 & 10.88 & 12.78 & 3.99 & 4.87 & 5.64 & 20.62 & 27.36 & {\ul 34.49} & 41.31 \\
\midrule
MTGNN & MAE & 2.70 & 3.06 & 3.51 & {\ul 1.34} & {\ul 1.67} & {\ul 1.97} & 12.88 & 18.96 & 23.32 & 28.52 \\
(2020) & RMSE & 5.20 & 6.19 & 7.26 & {\ul 2.82} & {\ul 3.78} & {\ul 4.51} & 23.37 & 32.48 & 38.35 & 45.24 \\
 & MAPE & 6.85 & 8.21 & {\ul 9.88} & {\ul 2.80} & {\ul 3.74} & {\ul 4.56} & {\ul 18.87} & 27.95 & 35.86 & 47.52 \\
 \bottomrule
\end{tabular}}

\end{table}

\begin{table}[]
\centering
\caption{Results compared with SOTA STGNNs on the META-LA dataset.}
\label{tab:stgnns}

\resizebox{\linewidth}{!}{
\begin{tabular}{cc|ccc|ccc|ccc}
\toprule
\multicolumn{2}{c|}{\multirow{2}{*}{Method}} & \multicolumn{3}{c|}{3} & \multicolumn{3}{c|}{6} & \multicolumn{3}{c|}{12} \\
\cmidrule{3-11}
\multicolumn{2}{c|}{} & MAE & RMSE & MAPE & MAE & RMSE & MAPE & MAE & RMSE & MAPE \\
\midrule
\multicolumn{2}{c|}{ST-Hyper} & \textbf{2.49} & \textbf{4.90} & \textbf{6.34} & \textbf{2.86} & \textbf{5.95} & \textbf{7.77} & \textbf{3.27} & \textbf{7.08} & \textbf{9.37} \\
\midrule
\multirow{3}{*}{High-order} 
 & MvHSTM$^*$(2025) & 2.62 & 5.03 & 6.72 & 2.96 & {\ul 6.00} & 8.11 & {\ul 3.40} & {\ul 7.15} & 9.95 \\
 & ASTHGCN$^*$(2023) & 2.65 & 5.05 & 6.80 & 3.01 & 6.07 & 8.26 & {\ul 3.40} & \textbf{7.08} & {\ul 9.81} \\
 & MixRNN$^*$(2023) & 2.63 & 5.06 & / & 3.00 & 6.08 & /  & 3.42 & 7.16 & / \\
 \midrule
\multirow{3}{*}{Adaptive} & MegaCRN$^*$(2023) & {\ul 2.52} & {\ul 4.94} & {\ul 6.44} & {\ul 2.93} & 6.06 & {\ul 7.96} & 3.38 & 7.23 & 9.72 \\
& PDFormer$^*$(2023) & 2.83 & 5.45 & 7.77 & 3.20 & 6.46 & 9.19 & 3.62 & 7.47 & 10.91 \\
& STAFormer$^*$(2023) & 2.65 & 5.11 & 6.85 & 2.97 & 6.00 & 8.13 & 3.34 & 7.02 & 9.70 \\
\midrule
\multirow{2}{*}{Hierachical} &  HiGP$^*$(2024) & 2.68 & 5.19 & 6.82 & 3.02 & 6.31 & 8.34 & {\ul 3.40} & 7.49 & 10.025 \\
 & HGCN(2021) & 2.89 & 5.47 & 7.44 & 3.34 & 6.60 & 8.92 & 3.87 & 7.82 & 10.51 \\
 \bottomrule
\end{tabular}
}

\end{table}

(3) \wu{As shown in Table \ref{tab:stgnns}, ST-Hyper outperforms SOTA STGNNs with adaptive, hierarchical, or high-order graph designs, achieving an average MAE reduction of 2.5\%. This improvement stems from its constrained hypergraph construction, where nodes (features at multiple spatio-temporal scales) are guided by a graph pooling loss and hyperedges are kept sparse, enhancing the robustness of learned dependencies.}

\begin{table}[]
\centering
\caption{Results of ablation study on METR-LA and China-AQI datasets.}
\label{tab:ablation}

\resizebox{\linewidth}{!}{
\setlength{\tabcolsep}{2pt}
{\fontsize{10}{10}\selectfont
\begin{tabular}{@{}c|cccccc|cccc|cc@{}}
\toprule
Variants  & \multicolumn{2}{c}{w/o STPM} & \multicolumn{2}{c}{ST-Hyper-M} & \multicolumn{2}{c|}{w/o $\mathcal{L}_{\text{GP}}$} & \multicolumn{2}{c}{w/o AHM} & \multicolumn{2}{c|}{ST-Hyper-H} & \multicolumn{2}{c}{ST-Hyper}    \\ \midrule
Metrics   & MAE           & RMSE         & MAE            & RMSE          & MAE                      & RMSE                    & MAE          & RMSE         & MAE            & RMSE          & MAE            & RMSE           \\ \midrule
METR-LA   & 2.96          & 6.12         & 2.93           & 6.07          & 2.84                     & 5.94                    & 2.92         & 6.05         & 2.87           & 6.01          & \textbf{2.81}  & \textbf{5.89}  \\ \midrule
China-AQI & 19.68         & 34.88        & 19.39          & 34.12         & 19.02                    & 33.24                   & 19.33        & 33.95        & 19.05          & 33.14         & \textbf{18.54} & \textbf{31.32} \\ \bottomrule
\end{tabular}
}
}

\end{table}

\subsection{Ablation Study}
\noindent\textbf{STPM module.} We evaluate three variants: \textit{1) w/o STPM}: removing the STPM module; \textit{2) ST-Hyper-M}: replacing memory-network-based graph learning with a non-memory method \cite{agcrn}, i.e., $\boldsymbol{A}=\operatorname{SoftMax}(\operatorname{ReLU}(\boldsymbol{E}_{1}\boldsymbol{E}_{2}^{\text{T}}))$; \textit{3) w/o $\mathcal{L}_{\text{GP}}$}: removing the graph pooling loss. Table \ref{tab:ablation} shows ST-Hyper outperforms all variants, highlighting the effectiveness of multi-ST-scale feature extraction, memory-network graph learning, and SPG regularization.

\noindent\textbf{AHM module.} We evaluate two variants: \textit{1) w/o AHM}: removing the AHM module; \textit{2) ST-Hyper-H}: replacing tri-phase hypergraph propagation with traditional hypergraph convolution \cite{hypergraphbai}. Table \ref{tab:ablation} shows ST-Hyper outperforms both, indicating that AHM’s tri-phase propagation effectively models high-order dependencies across multiple ST-scales.

\begin{table}[]
\centering
\caption{MSE Results with different noise intensities on the Solar-Energy dataset.}
\label{tab:noise}

\resizebox{0.6 \linewidth}{!}{
\begin{tabular}{cccccc}
\toprule
SNR & 100db & 80db & 60db & 40db \\
\midrule
ST-Hyper(Ours) & \textbf{0.185} & \textbf{0.190} & \textbf{0.202} & \textbf{0.241}\\
iTransformer & 0.203 & 0.228 & 0.240 & {\ul 0.302}\\
TimeMixer & {\ul 0.195} & {\ul 0.206} & {\ul 0.232} & 0.320 \\
CrossGNN & 0.297 & 0.308 & 0.336 & 0.401 \\
\bottomrule
\end{tabular}}
\end{table}

\subsection{Robustness Analysis}
\wu{To evaluate model robustness under noisy conditions, we follow the protocol of CrossGNN~\cite{crossGNN} and add Gaussian white noise of varying intensities to the Solar-Energy dataset. We fix input and output lengths to 96 and use MSE as the evaluation metric. As shown in Table~\ref{tab:noise}, we progressively decrease the signal-to-noise ratio (SNR) from 100 dB to 40 dB, which introduces increasing levels of noise. Across all noise levels, ST-Hyper consistently outperforms the baselines, with relative improvements in MSE ranging from 5.12\% to 20.20\%.
These results highlight the strong noise resilience of ST-Hyper. We attribute this robustness to its ability to capture and integrate patterns across multiple spatio-temporal scales, allowing representations at different granularities to reinforce each other and mitigate the impact of local anomalies or perturbations.}

\subsection{Hyperparameter Sensitivity Study}

\begin{figure}[]
    \centering
    \includegraphics[width= 0.47\textwidth]{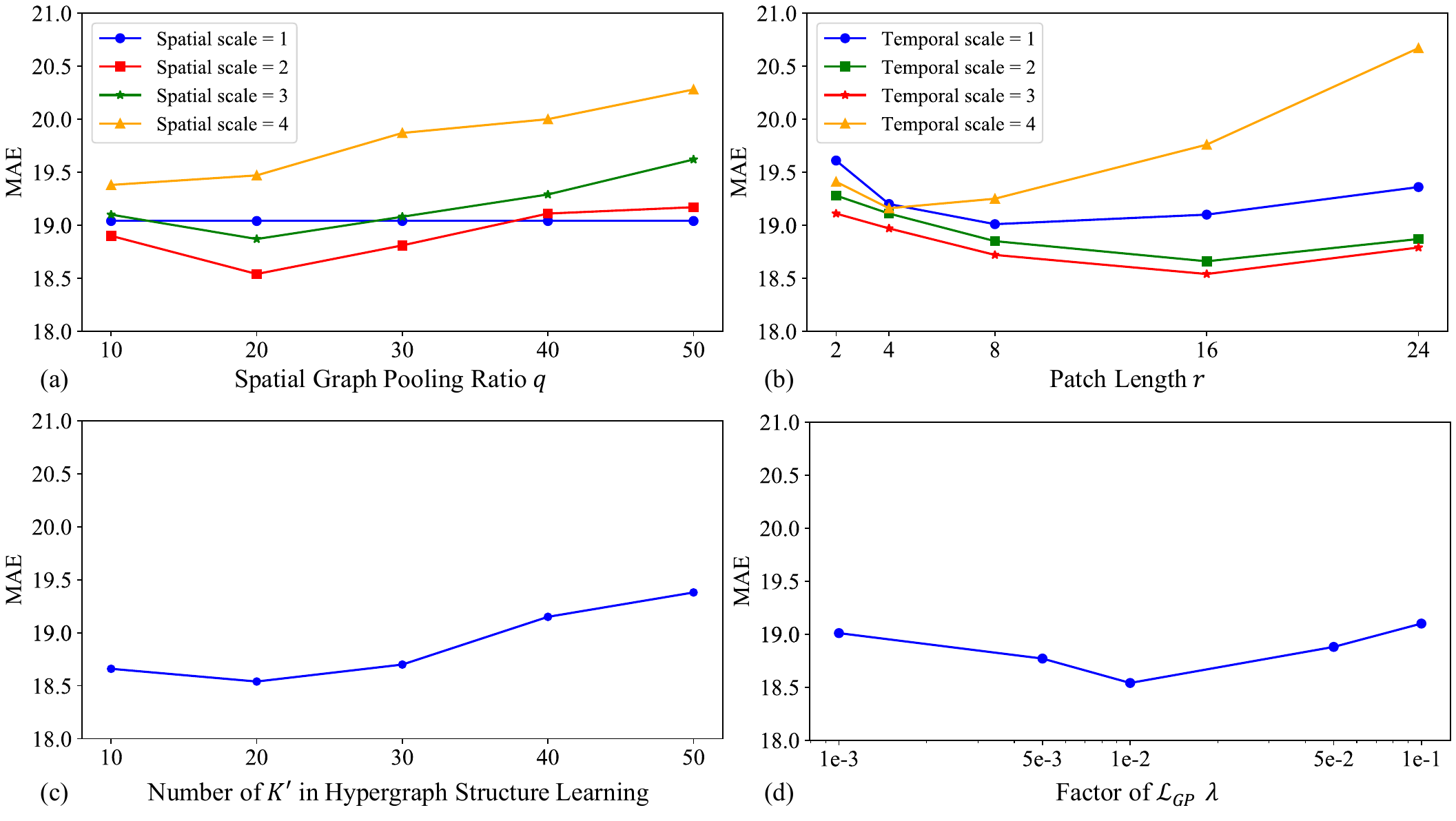}
    
    \caption{Results of hyperparameter sensitivity study on China-AQI dataset.}
    
    \label{fig:hyper}
\end{figure}

\noindent \textbf{Number of Spatial Scales (\# Spatial Scale) and Spatial Graph Pooling Ratio $ q$.} 
These two hyperparameters determine the SPG structure. $q$ determines the number of nodes in the spatial graph at spatial scale $j$ (i.e., $N_{j} = \left\lfloor{{N_{j-1}}/{q}}\right\rfloor$). As shown in Figure \ref{fig:hyper}(a), the optimal \# Spatial Scale is 2 and the optimal $q$ is 20. The reason is that a smaller \# Spatial Scale may fail to model dependencies at multiple spatial scales, while a larger \# Spatial Scale may introduce too many parameters and make the model difficult to converge. In addition, a smaller $q$ may cause irregular nodes to be grouped into a group node, while a larger $q$ results in too few nodes at larger spatial scales and limits model capacity.

\noindent \textbf{Number of Temporal Scales (\# Temporal Scale) and Patch Length $ r$.} 
These two hyperparameters determine how features at multiple temporal scales are extracted for each spatial scale. As shown in Figure \ref{fig:hyper}(b), the optimal \# Temporal Scale is 3 and the optimal $r$ is 16. The reason is that a smaller \# Temporal Scale may fail to model long-term dependencies, while a larger \# Temporal Scale may introduce too many parameters. In addition, a smaller $r$ may overly focus on fine-grained noise, while a larger $r$ may overlook important short-term patterns.

\noindent \textbf{Number of $ K'$.} $ K'$ determines the number of nodes connected by each hyperedge in the hypergraph. As shown in Figure \ref{fig:hyper}(c), the optimal $K'$ is 20. The reason is that a smaller $K'$ may limit the ability of ST-Hyper to model group-wise dependencies among nodes, while a larger $K'$ may introduce noise into hyperedge features. 

\noindent \textbf{Factor of} $\mathcal L_{\text {GP}}$ $\lambda$\textbf{.} $\lambda$ determines the contribution of graph pooling loss $\mathcal L_{\text {GP}}$ to training loss $\mathcal L_{\text {train}}$. As shown in Figure \ref{fig:hyper}(d), the optimal $\lambda$ is 1e-2. The reason is that a smaller $\lambda$ may lead to insufficient regularization to SPG learning, while a larger $\lambda$ may overly constrain the model.

\begin{figure}
    \centering
    \includegraphics[width=0.7\linewidth]{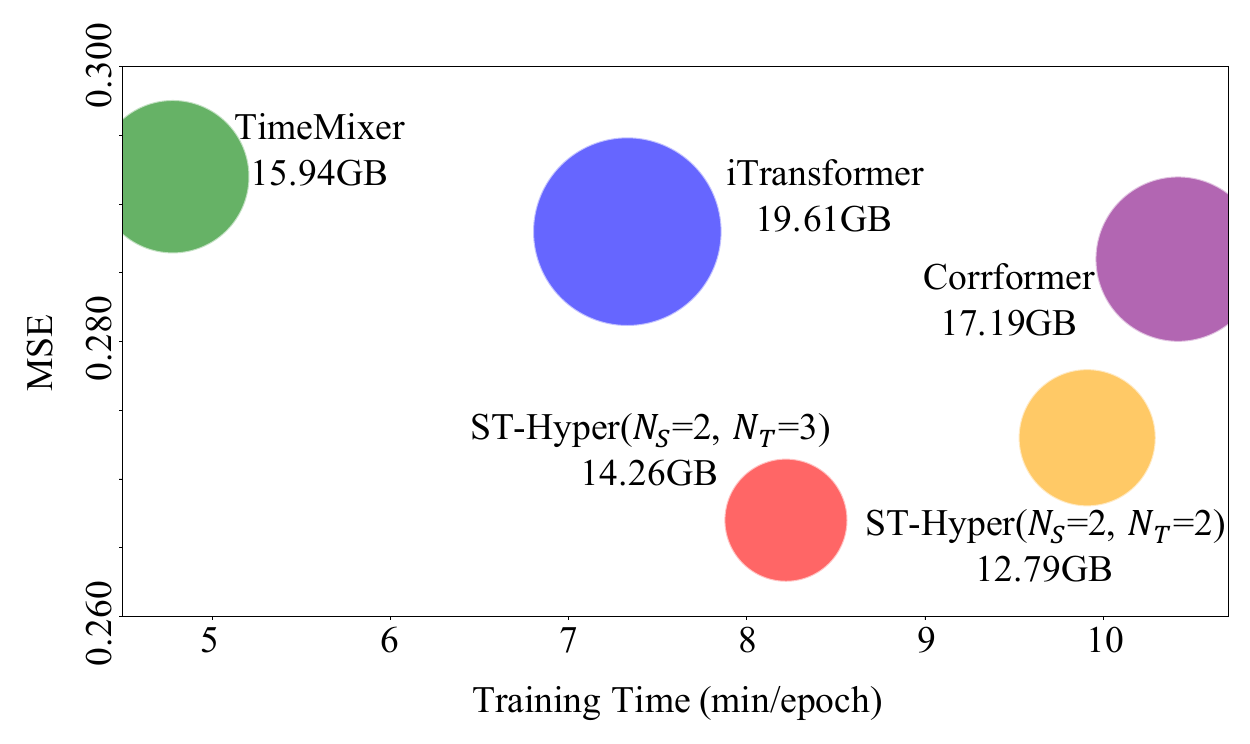}
    \caption{Results of computation cost comparison, where $N_S$ and $N_T$ indicate the number of spatial and temporal scales, respectively, for simplicity.}
    \label{fig:efficiency}
\end{figure}

\begin{figure*}[]
\centering
\includegraphics[width=1\textwidth]{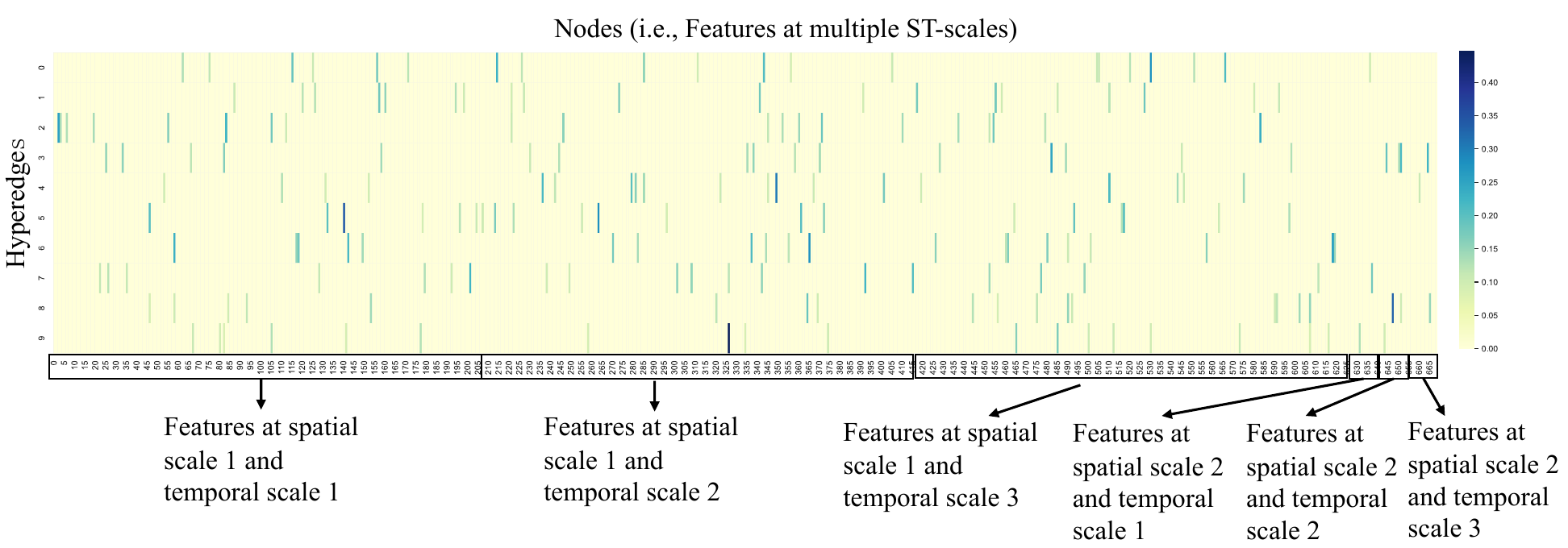}
\caption{Visualization of sparsified hypergraph incidence matrix on China-AQI dataset.}
\label{figure_vis_2}
\end{figure*}

\begin{figure}[]
\centering
\includegraphics[width=0.4 \textwidth]{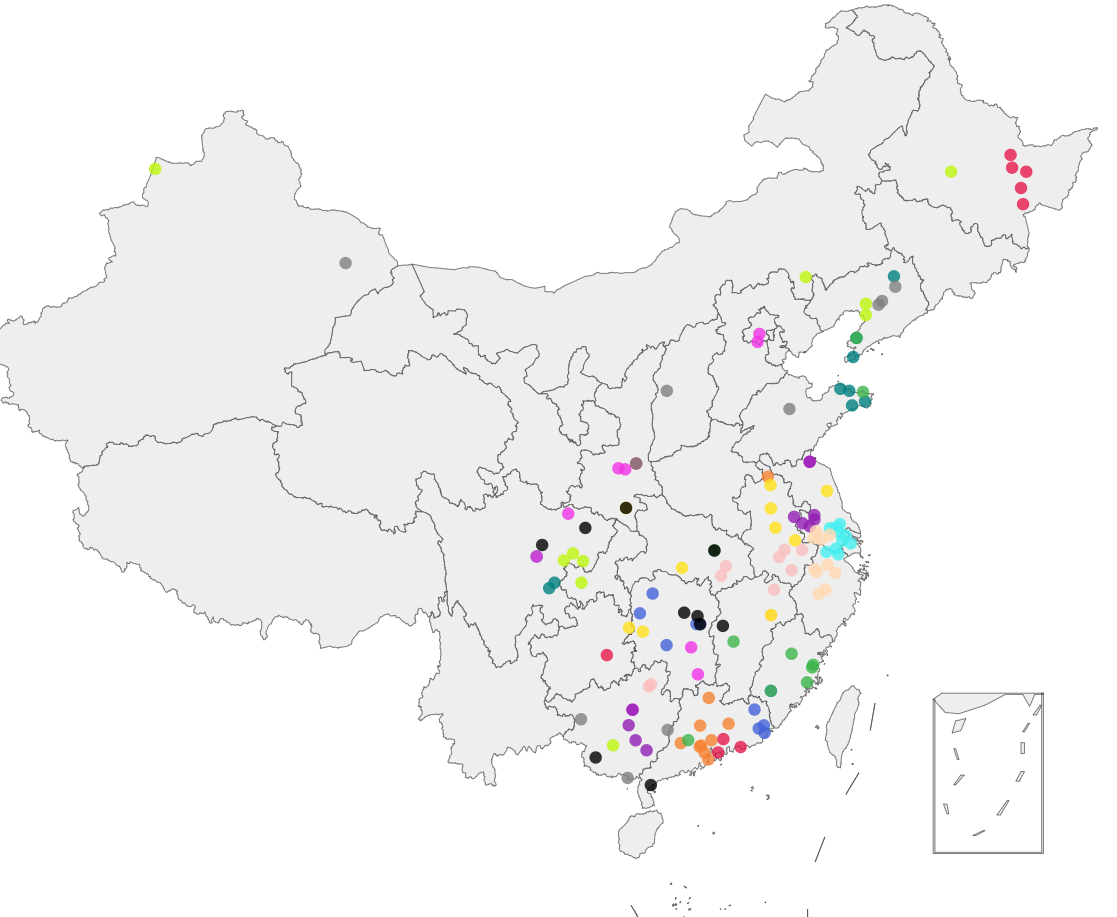}
\caption{Visualization of the grouping result of Spatial Pyramidal Graph Learning with 2 spatial scales. Different colors represent different group nodes at spatial scale 2.}
\label{figure_vis_1}
\end{figure}

\begin{figure}[]
    \centering
    \includegraphics[width= 0.47\textwidth]{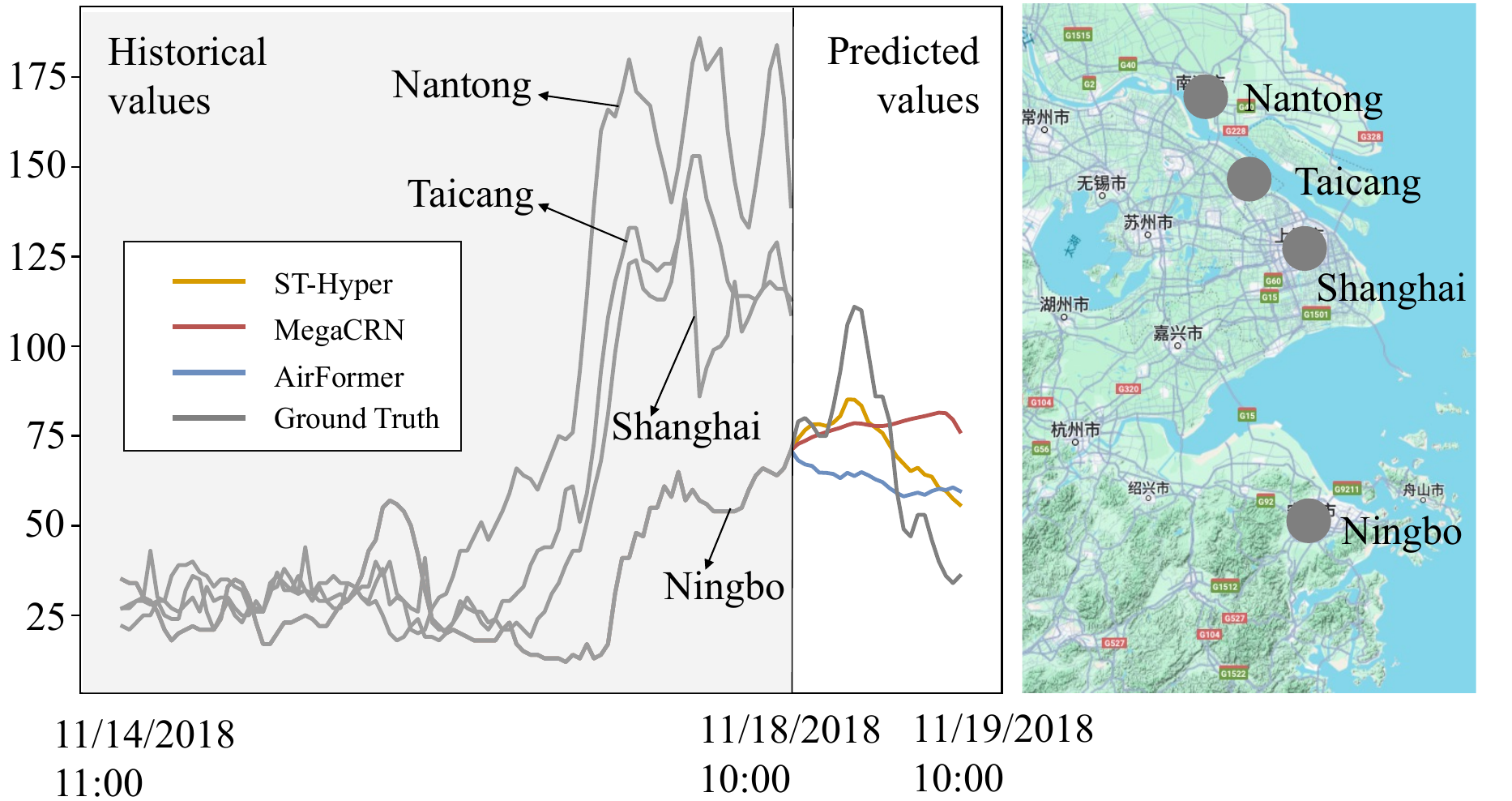}
    \caption{Visualization of predicting AQI for Ningbo city.}
    \label{fig:case}
\end{figure}

\subsection{Computational Cost Comparison}

We evaluate ST-Hyper against competitive baselines in training time, GPU memory, and forecasting accuracy, following iTransformer \cite{itransformer}, on the Temperature dataset with 3,850 variables. 
As shown in Fig. \ref{fig:efficiency}, ST-Hyper achieves the highest accuracy while using the least GPU memory, thanks to its ability to capture high-order dependencies across multiple spatiotemporal scales using sparse hypergraphs. 
ST-Hyper requires longer training than TimeMixer, which ignores spatial dependencies, and iTransformer, which ignores multi-scale dynamics. 
Compared with Corrformer, which models spatial and temporal dependencies independently, ST-Hyper demonstrates superior computational efficiency and predictive accuracy. 
Overall, ST-Hyper delivers SOTA performance with favorable computational cost, making it practical for real-world forecasting.

\subsection{Visualization}
\noindent \textbf{Sparsified Hypergraph Incidence Matrix.}
Figure \ref{figure_vis_2} portrays the subpart of the sparsified hypergraph incidence matrix, i.e., $\tilde{\bm \Lambda}$, on China-AQI dataset. Recall that the entries in the sparsified hypergraph incidence matrix represent assigning possibilities, ranging from 0 to 1, from nodes to hyperedges. The nodes in the hypergraph represent features from multiple ST-scales. On China-AQI dataset, the number of spatial scales is set to 2 and the number of temporal scales is set to 3. As shown in Figure \ref{figure_vis_2}, features at different ST-scales are associated with a certain hyperedge. In addition, the member nodes for different hyperedges are not identical, reflecting the diversity of hyperedge information. AHM module incorporating tri-phase hypergraph propagation models the interaction between hyperedges, which can model the high-order dependencies across multiple ST-scales. 

\noindent \textbf{Spatial Pyramidal Graph Structure.}
The visualization in Figure \ref{figure_vis_1} depicts the grouping result for variables on China-AQI dataset, with 2 spatial scales. For each group node at spatial scale 2, we present the nodes at spatial scale 1, with top 10 possibilities, that are assigned to the group node. As shown in Figure \ref{figure_vis_1}, we note a significant adjacency among nodes within one group node, primarily owing to their geographical proximity. In addition, there are instances where nodes, despite being geographically distant, are grouped together. This illustrates that our method ST-Hyper may consider a broader range of spatial features when modeling spatial dependencies.

\noindent \textbf{Predicted AQI.}
We select a case that predicts the future AQI values for Ningbo city on China-AQI dataset to illustrate the superiority of ST-Hyper. As shown in Figure \ref{fig:case}, the left part is the historical values (96 hours) of Ningbo city and its neighboring cities and the predicted values (24 hours) of different methods. The right part portrays the geometric distribution of these cities. During this period, a sharp increase occurs in Ningbo city, making forecasting challenging. While the SOTA STGNN-based (MegaCRN) and Transformer-based methods (AirFormer) fail to capture this increase, ST-Hyper successfully does so. This is because ST-Hyper can detect sudden changes in nearby cities in the past by extracting features at multiple ST-scales (e.g., city-hour and city group-day scales), and pass the information about sudden changes to the target city (Ningbo) through interacting features from multiple ST-scales.

\section{Conclusions}
In this work, we propose ST-Hyper for MTS forecasting, which can model high-order dependencies across multiple ST-scales through adaptive hypergraph modeling. 
Extensive experiments on six real-world datasets demonstrate that ST-Hyper achieves SOTA performance, outperforming the best baseline with an average MAE reduction of 3.8\% and 6.8\% for long-term and short-term forecasting, respectively. \wu{Furthermore, additional analyses underscore the robustness of ST-Hyper, as well as the practical importance of modeling dependencies across multiple ST-scales in real-world scenarios.
In future work, we will extend ST-Hyper in several directions, including: (1) discovering dynamic hypergraph evolution to capture time-varying high-order dependencies, (2) uncovering directed hypergraphs to enhance interpretability, and (3) improving efficiency to scale ST-Hyper to ultra-large datasets with high-dimensional inputs.}

\bibliographystyle{ACM-Reference-Format}
\bibliography{reference}

\end{document}